\documentclass[letterpaper, 10 pt, conference]{ieeeconf}  

\IEEEoverridecommandlockouts                              

\usepackage{graphicx} 
\usepackage{amsmath} 
\usepackage{amssymb}  
\usepackage{multirow}
\usepackage{booktabs}
\usepackage{threeparttable}
\usepackage{cite}

\title{\LARGE \bf
Biomimetic Control of Myoelectric Prosthetic Hand Based on a Lambda-type Muscle Model
}

\author{Akira Furui, Kosuke Nakagaki, and Toshio Tsuji
\thanks{Corresponding authors: Akira Furui and Toshio Tsuji.}
\thanks{A. Furui and T. Tusji are with the Graduate School of Advanced Science and Engineering,
        Hiroshima University, Higashi-hiroshima, 739-8527 Japan (e-mail: akirafurui@hiroshima-u.ac.jp; tsuji@bsys.hiroshima-u.ac.jp)
        }%
\thanks{K. Nakagaki is with the Graduate School of Engineering, Hiroshima University, Higashi-hiroshima, 739-8527 Japan (e-mail: nakagaki@bsys.hiroshima-u.ac.jp}%
\thanks{Copyright (c) 2021 IEEE.  Personal use of this material is permitted.  Permission from IEEE must be obtained for all other uses, in any current or future media, including reprinting/republishing this material for advertising or promotional purposes, creating new collective works, for resale or redistribution to servers or lists, or reuse of any copyrighted component of this work in other works. 
This paper is accepted for publication at the International Conference on Robotics and Automation 2021 (ICRA 2021).}
}

\begin{document}

\maketitle
\thispagestyle{empty}
\pagestyle{empty}

\begin{abstract} 
  Myoelectric prosthetic hands are intended to replace the function of the amputee's lost arm.
  Therefore, developing robotic prosthetics that can mimic not only the appearance and functionality of humans but also characteristics unique to human movements is paramount.
  Although the impedance model was proposed to realize biomimetic control, this model cannot replicate the characteristics of human movements effectively because the joint angle always converges to the equilibrium position during muscle relaxation.
  This paper proposes a novel biomimetic control method for myoelectric prosthetic hands integrating the impedance model with the concept of the $\lambda$-type muscle model.
  The proposed method can dynamically control the joint equilibrium position, according to the state of the muscle, and can maintain the joint angle naturally during muscle relaxation.
  The effectiveness of the proposed method is evaluated through simulations and a series of experiments on non-amputee participants.
  The experimental results, based on comparison with the actual human joint angles, suggest that the proposed method has a better correlation with the actual human motion than the conventional methods.
  Additionally, the control experiments showed that the proposed method could achieve a natural prosthetic hand movement similar to that of a human, thereby allowing voluntary hand opening and closing movements.
\end{abstract}

\vspace{2mm}

\section{Introduction}

A study published by the cabinet office of Japan reported more than 82,000 upper-limb amputations due to accidents or disease~\cite{noauthor_2013-ve}. 
As a part of life support for upper-limb amputees, a myoelectric prosthetic hand is prescribed.
The myoelectric prosthetic hand is controlled by inferring the amputee's intentions from electromyogram (EMG), which is the electrical activity produced by muscles.
As the myoelectric prosthetic hand can be manipulated voluntarily according to the muscle force exerted, there is a possibility that users would operate it as they use their hands.
Hence, various related research and development have been developed in this regard.

Myoelectric prosthetic hands are designed to replace the function of the upper limb by replacing the lost arm.
Thus, the prosthetic must realize human-like behavior that mimics the movements of the human hand in addition to its appearance and operability.
MyoBock~\cite{OttoBock_undated-wi} is currently the most popular myoelectric prosthetic hand and enables control of two hand movements (grasp and open).
Its appearance is improved by wearing a glove that resembles a human arm.
In addition, the Vincent evolution 3~\cite{Vincent_Systems_undated-fu} controls multiple fingers independently, making it possible to reproduce various types of human gripping motions.
However, most myoelectric prosthetic hands studied and developed thus far use a simple proportional control to open and close the hand in response to the muscle strength~\cite{OttoBock_undated-wi,Vincent_Systems_undated-fu,Parker2006-uj,Atzori2015-ha,Wang2017-qq}.
Therefore, human motion characteristics have not been sufficiently considered for these myoelectric prosthetic hands, and human-like behavior has not been realized entirely.

Studies on biomimetic control of prosthetic hands have focused on reflecting the human motion characteristics in the prosthetic hand~\cite{Kakoty2014-mo,Furui2019-bz,Laffranchi2020-vp}.
The authors previously introduced the impedance model-based biomimetic control method~\cite{Tsuj2000-ho,Tsuji2010-cg} to control a prosthetic hand~\cite{Furui2019-bz}.
This model can realize natural human-like hand movements by controlling the actuators, considering the inertia, viscosity, and stiffness around the human joint.
We showed that smooth movement according to the user's EMG signals could be achieved through the experiments conducted on intact and amputee participants~\cite{Furui2019-bz}.
However, in the impedance control, the equilibrium angle at the resting state is fixed at the origin as the equilibrium position of the spring, and in consequence, the joint angle always converges to the origin, regardless of its position.
In contrast, if the actual human movement is considered, it is possible to maintain the joint angle during muscle relaxation after realizing the desired joint angle by muscle contraction.
Therefore, the previously proposed impedance control is insufficient to replicate the characteristics of human movement.

For this problem, if we could develop a control system considering the movement characteristics of the muscles, it would be possible to control the equilibrium position of the joint angle dynamically, according to the contraction state of the muscles.
Various theories have been proposed to model muscle dynamics~\cite{Merton1953-ll, Bizzi1982-xa, Feldman1986-va}. 
Merton proposed the servo hypothesis ($\gamma$-model), assuming that the tonic stretch reflex is the primary motor control mechanism~\cite{Merton1953-ll}.
Bizzi \textit{et al.} proposed an $\alpha$-model~\cite{Bizzi1982-xa}, relying on the hypothesis that the brain mainly controls the activity of $\alpha$ motor neurons.
Moreover, Feldman proposed the $\lambda$-model~\cite{Feldman1986-va}, assuming that the command from the central nervous system controls the threshold of the tonic stretch reflex.
This model assumes that the muscle contracts when the muscle length is longer than the threshold, and conversely, when it is shorter, the muscle activity stops.
The $\lambda$-model is prone to be superior to other models in that it can explain the entire voluntary movement of humans to some extent in a concise theory~\cite{Latash2010-lm}.
Thus, the theory has been used as a major motion control theory for several decades.

This paper proposes a biomimetic control method for myoelectric prosthetic hands based on the impedance property of muscle and the concept of the $\lambda$-model.
The proposed method can reflect human motor characteristics to the movement of the prosthetic hand by dynamically controlling the equilibrium angle of the joint according to the state of contraction/relaxation of the muscle.
In the experiments, the effectiveness of the proposed method is verified through simulations and control experiments.

\section{Biomimetic Control Based on the $\lambda$-type Muscle Model}
\subsection{Impedance Control Model}
\begin{figure}[t]
  \centering
  \includegraphics[width=0.73\hsize]{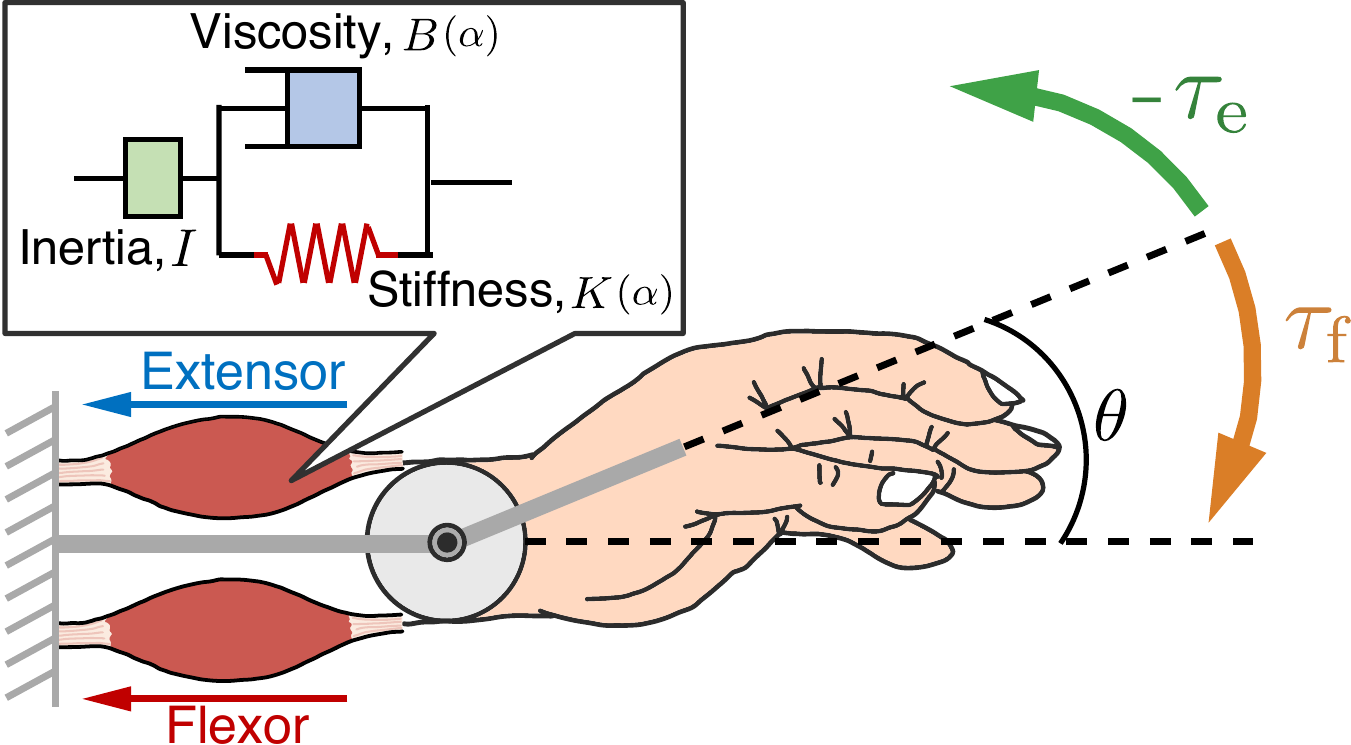}
  \caption{Biomimetic impedance control based on the musculoskeletal model of human joint.
  (a) The characteristics of joint motion can be represented by the tension balance between flexors and extensors.
  By reflecting the impedance characteristics of human joints, natural and smooth control of the prosthetic hand can be achieved.
  }
  \label{fig:concept}
\end{figure}

First, based on impedance control~\cite{Tsuji2010-cg}, the characteristics of joint motion is represented by the tension balance between flexors and extensors (Fig.~\ref{fig:concept}).
The equation of motion around each joint $j~(j=1,~2,~\dots,~J)$ of the prosthetic hand is defined as
\begin{eqnarray}
  I_{j}\ddot{\theta_{j}}+B_{j}(\alpha_{j})\dot{\theta_{j}}+K_{j}(\alpha_{j})(\theta_{j}-\theta^{0}_{j}) = \tau_{j \mathrm{f}} - \tau_{j \mathrm{e}},
  \label{motion}
\end{eqnarray}
where $I_{j}$, $B_{j}(\alpha_{j})$, and $K_{j}(\alpha_{j})$ are the inertia, viscosity, and stiffness, respectively.
$\theta_{j}$ and $\theta^{0}_{j}$ are the joint angle and its equilibrium angle for stiffness element only.
Furthermore, $\tau_{j \mathrm{f}}$ and $\tau_{j \mathrm{e}}$ are the joint torque generated by the flexor and extensor, respectively.
The change in the characteristics associated with muscle activity is expressed by defining the stiffness and viscosity as functions of the muscle contraction level $\alpha_j$ $(0 \leq \alpha_j \leq 1)$.
\begin{align}
  K_j(\alpha_j) &= k_{j,1}\alpha_j^{k_{j,2}}+k_{j,3},\\
  B_j(\alpha_j) &= b_{j,1}\alpha_j^{b_{j,2}}+b_{j,3},
\end{align}
The equilibrium angle $\theta_j^{\mathrm{eq}}$ of the complete system can be calculated as the joint angle $\theta_j$ when $\ddot{\theta}$ and $\dot{\theta}=0$.
\begin{equation}
  \theta_j^{\mathrm{eq}} = \frac{1}{K_j(\alpha_j)} (\tau_{j \mathrm{f}} - \tau_{j \mathrm{e}}) + \theta_j^0
  \label{eq:theta_eq}
\end{equation}

\subsection{Model Requirements}
From (\ref{eq:theta_eq}), the torque $\tau_j(t)$ and the equilibrium angle $\theta_0$ of the joint angle can be varied adaptability to control the equilibrium angle of the model dynamically, according to the muscle contraction state.
This study designs these variables to build a control model satisfying the following specifications.
\begin{itemize}
  \item \textbf{Specification 1}: When the muscle relaxes, the equilibrium position is maintained at the value just before relaxation.
  \item \textbf{Specification 2}: Continuity of the equilibrium position is maintained even when the direction of motion changes from flexion to extension or extension to flexion.
\end{itemize}

When designing the model, it is desirable to consider all the muscle information for the joint movement.
However, various muscles act when a person performs a complicated operation, such as flexing their fingers, and it is thus difficult to estimate all the muscle information on the joint motion from the EMG signals.
Therefore, we switch the muscle contraction level depending on the classified motion as follows:
\begin{equation}
  \alpha_j(t) = \begin{cases}
    \alpha_{j\mathrm{f}}(t) & (\mathrm{Flexion\ motion})   \\
    \alpha_{j\mathrm{e}}(t) & (\mathrm{Extension\ motion})
  \end{cases}.
\end{equation}
Furthermore, the torque $\tau_{j\rm{f}},~\tau_{j\rm{e}}$ can be expressed as a function of the muscle contraction level $\alpha(t)$:
\begin{eqnarray}
  \tau_{ji}(t)=\alpha_{ji}(t) \tau_{ji}^{\rm{max}},
  \label{tau}
\end{eqnarray}
where $\tau_{ji}^{\rm{max}}$ is the maximum value of each predetermined torque, and the subscript $i \in \{\mathrm{f}, \mathrm{e}\}$ indicates flexor and extensor.
Henceforth, we omit $(t)$ to indicate a function of time, except when highlighting the temporal context.

\subsection{Invariant Condition During Muscle Relaxation}
\begin{figure}[t]
  \centering
  \includegraphics[width=0.75\hsize]{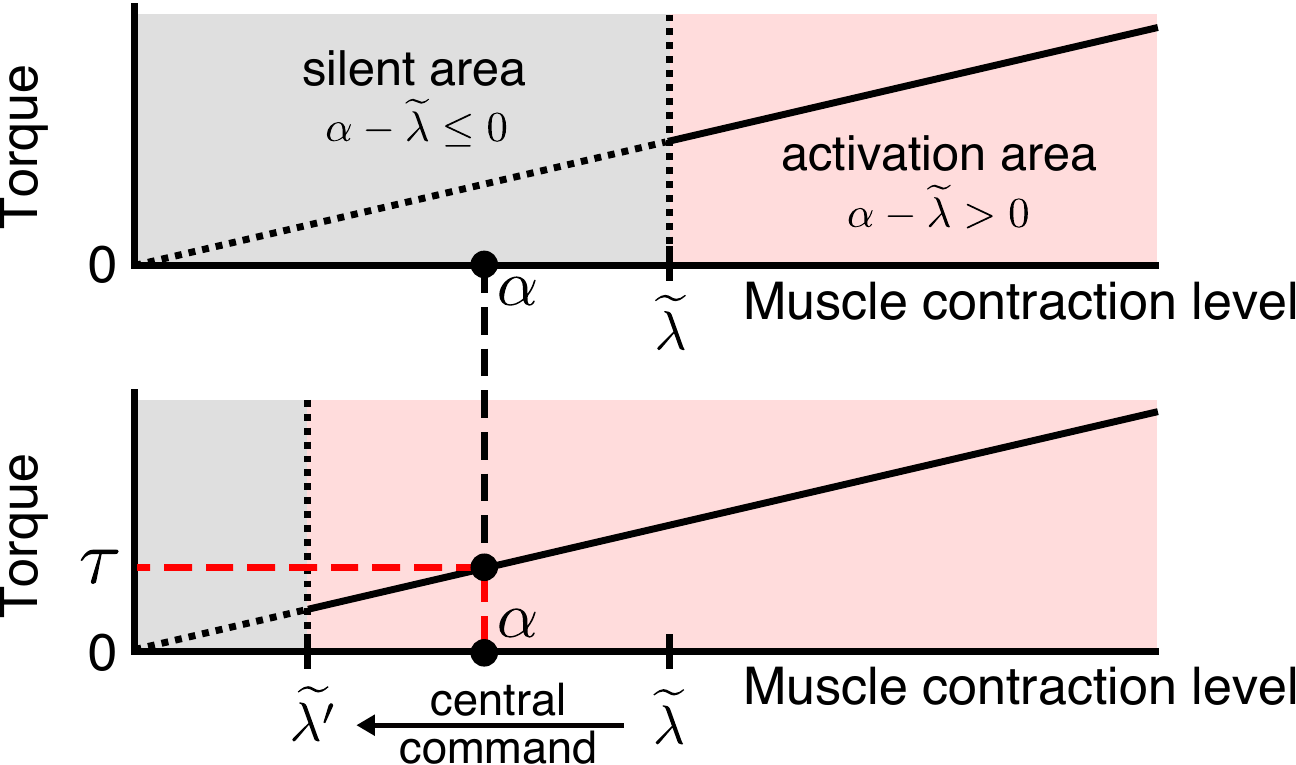}
  \caption{Invariant condition during muscle relaxation based on the concept of the $\lambda$-model.
  The silent and activation areas are provided by the muscle contraction level threshold $\widetilde{\lambda}$.
  The torque is generated only when the current muscle contraction level is in the activation area.
  }
  \label{fig:concept2}
\end{figure}

We first design a model satisfying the specification 1.
In the model design, we adopt the concept of the $\lambda$-model~\cite{Feldman1986-va}, which is a model of human muscle movements.
The $\lambda$-model assumes that the brain achieves motion by controlling the muscle length $x$, while the intensity of the motor commands determines the threshold $\lambda$ of the muscle length.
Motion conditions of the muscle are formulated as follows:
\begin{eqnarray}
  x - \lambda > 0. \label{feldman}
\end{eqnarray}
The muscle is active if the condition of (\ref{feldman}) expression is satisfied; otherwise, the muscle activity is stopped, and the current state is maintained.
Here, the muscle activity level $s$ in the $\lambda$-model is defined as a function of the muscle length $x$ and threshold $\lambda$.
Moreover, the muscle activity level is assumed to be a simple linear function that can be written as follows:
\begin{eqnarray}
  s=s(x,~\lambda)=G(x-\lambda)+h,\label{s}
\end{eqnarray}
where $G$ and $h$ are constants.
Equation (\ref{s}) can be rewritten as
\begin{eqnarray}
  x-\lambda=\frac{1}{G}(s-h).\label{s2}
\end{eqnarray}
Here, we assume that the constant $G$ is the maximum value $s^{\rm{max}}$ of $s$, and $s/G$ in (\ref{s2}) can be regarded as the muscle contraction level $\alpha_{ji}$.
Furthermore, if the constant $h/G$ is regarded as the threshold value $\widetilde{\lambda}_{ji}$ of the muscle contraction level, from the conditional expression (\ref{feldman}), the muscle length $x$ can be replaced by the muscle contraction level $\alpha_{ji}$, and the muscle length threshold $\lambda$ can be replaced by the muscle contraction level threshold $\widetilde{\lambda}_{ji}$.
Subsequently, the muscle activity condition in the control model is defined as follows:
\begin{eqnarray}
  \alpha_{ji} - \widetilde{\lambda}_{ji} > 0. \label{feldman2}
\end{eqnarray}
Here, we introduce a function expressing the active state of the muscle using the activity condition (\ref{feldman2}) as follows:
\begin{eqnarray}
  C^{+} = [\alpha_{ji} - \widetilde{\lambda}_{ji}]^{+} =
  \begin{cases}
    1 & (\alpha_{ji} - \widetilde{\lambda}_{ji} > 0)    \\
    0 & (\alpha_{ji} - \widetilde{\lambda}_{ji} \leq 0)
  \end{cases}
  ,
  \label{c+}
\end{eqnarray}
\begin{equation}
  C^{-} = 1 - C^{+},
  \label{c-}
\end{equation}
where $C^{+}$ is a function expressing the active state of the muscle and becomes $C_{i}^{+} = 1$ when $\alpha_{ji}$ is larger than the threshold $\widetilde{\lambda}_{ji}$.
Meanwhile, $C_{i}^{-}$ is a function expressing the inactive state of the muscle and becomes $C_{i}^{-}=1$ when $\alpha_{ji}$ is equal to or less than the threshold $\widetilde{\lambda}_{ji}$.
We consider that the joint angle varies when the muscular activity condition is satisfied and is maintained when the muscle is in an inactive state, according to the concept of the $\lambda$-model (Fig.~\ref{fig:concept2}).
Then, using (\ref{c+}) and (\ref{c-}), we define $\tau_{j\rm{f}},~\tau_{j\rm{e}}$, and $\theta^{0}_{ji}$ as follows:
\begin{eqnarray}
  \tau_{ji}&=&C^{+}_{i}\tau_{ji}^{\rm{max}}\alpha_{ji},
  \label{tauC}\\
  \theta^{0}_{ji}(t)&=&C_{i}^{-}\theta^{\mathrm{eq}}_{ji}(t-\Delta t),
  \label{tauCtheta}
\end{eqnarray}
where $\theta^{\mathrm{eq}}_{ji}(t-\Delta t)$ is the equilibrium angle before the sampling time $\Delta t$. 

Subsequently, we preform the control using muscle information only in the flexion direction or extension direction based on the constraint condition mentioned above. 
A separate model is defined for each direction, and control is achieved by switching between models.
Under this condition, when a model is applied in a specific direction, the muscle contraction in the other direction is zero.
Therefore, when the control is performed based on (\ref{tauC}) and (\ref{tauCtheta}), the torque and the equilibrium position are discontinuous before and after switching the models.
The continuity of the equilibrium position can be maintained by complementing the muscle contraction level in the other direction with the muscle contraction level in the classified direction.
In cases other than co-contraction, it is assumed that one muscle contraction level gradually decreases as the other muscle contraction level increases.
Based on this assumption, when the flexion direction is classified, the muscle contraction level $\alpha_{j\rm{e}}$ in the extension direction is expressed using the muscle contraction level $\alpha_{j\rm{f}}$ in the flexion direction:
\begin{eqnarray}
  \alpha_{j\rm{e}}=A_{\rm{e}}(1-\alpha_{j\rm{f}}), \label{fpre}
\end{eqnarray}
where $A_{\rm{e}}$ is a constant.
Here, substituting (\ref{fpre}) into (\ref{tauC}) and expressing torque in the flexion direction, the control model is
\begin{eqnarray}
  \begin{cases}
    \tau_{j\rm{f}}=C^{+}_{\rm{f}}\tau_{j\rm{f}}^{\rm{max}}\alpha_{j\rm{f}} \\
    \tau_{j\rm{e}}=C^{+}_{\rm{f}}\tau_{j\rm{e}}^{\rm{max}}A_{\rm{e}}(1-\alpha_{j\rm{f}})
    \label{tauF}
  \end{cases}
  \mbox{(Flexion direction)}.
\end{eqnarray}
Similarly, when the extension direction is classified, the muscle contraction level $\alpha_{j\rm{f}}$ in the flexion direction is expressed using the muscle contraction level $\alpha_{j\rm{e}}$ in the extension direction:
\begin{eqnarray}
  \alpha_{j\rm{f}}=A_{\rm{f}}(1-\alpha_{j\rm{e}}), \label{epre}
\end{eqnarray}
where $A_{\rm{f}}$ is a constant.
Thus, the control model in the extension direction is 
\begin{eqnarray}
  \begin{cases}
    \tau_{j\rm{f}}=C^{+}_{\rm{e}}\tau_{j\rm{f}}^{\rm{max}}A_{\rm{f}}(1-\alpha_{j\rm{e}}) \\
    \tau_{j\rm{e}}=C^{+}_{\rm{e}}\tau_{j\rm{e}}^{\rm{max}}\alpha_{j\rm{e}}
    \label{tauE}
  \end{cases}
  \mbox{(Extension direction)}.
\end{eqnarray}
From the above equations, considering the muscle activity condition based on the $\lambda$-model for muscle motion, we can construct a model that maintains the equilibrium position of the joint angle even when the muscle relaxes.

\subsection{Continuity Condition for Switching of the Motion Direction}
We next consider the continuity of the equilibrium position at the time of changing the model to satisfy the specification 2.
We consider that a model is applied in the extension direction at time $t_{\rm e}$ and then in the flexion direction at time $t_{\rm e}+\Delta t$.
At this time, the equation of motion of the model in the extension direction applied at $t_{\rm e}$ becomes
\begin{align}
  I_{j}\ddot{\theta_{j}} & +B_{j}(\alpha_{j\rm{e}}(t_{\rm{e}}))\dot{\theta_{j}}+K_{j}(\alpha_{j\rm{e}}(t_{\rm{e}}))\theta_{j} \nonumber                             \\
                         & =\tau_{j\rm{f}}^{\rm{max}}A_{\rm{f}}(t_{\rm{e}})(1-\alpha_{j\rm{e}}(t_{\rm{e}})) -\tau_{j\rm{e}}^{\rm{max}}\alpha_{j\rm{e}}(t_{\rm{e}}).
  \label{pren}
\end{align}
The equation of motion of the model in the flexion direction applied at $t_{\rm{e}}+\Delta t$ immediately thereafter is given as
\begin{align}
  I_{j}\ddot{\theta_{j}} & +B_{j}(\alpha_{j\rm{f}}(t_{\rm{e}}+\Delta t))\dot{\theta_{j}}+K_{j}(\alpha_{j\rm{f}}(t_{\rm{e}}+\Delta t))\theta_{j} \nonumber \\
                         & =\tau_{j\rm{f}}^{\rm{max}}\alpha_{j\rm{f}}(t_{\rm{e}}+\Delta t) \nonumber                                                      \\
                         & \quad -\tau_{j\rm{e}}^{\rm{max}}A_{\rm{e}}(t_{\rm{e}}+\Delta t)(1-\alpha_{j\rm{f}}(t_{\rm{e}}+\Delta t)).
  \label{pren+1}
\end{align}
The equilibrium positions of (\ref{pren}) and (\ref{pren+1}) must be equal to maintain continuity at the time the model switches.
Here, we assume that the state of motion is the steady state $(\ddot{\theta_{j}},~\dot\theta_{j}=0)$ just before switching the direction of motion. The equilibrium positions in (\ref{pren}) and (\ref{pren+1}) are
\begin{align}
  \theta_{j}(t_{\rm e})=\frac{1}{K_{j}(\alpha_{j\rm{e}}(t_{\rm{e}}))} \{ & \tau_{j{\rm f}}^{\rm max}A_{\rm f}(t_{\rm e})(1-\alpha_{j{\rm e}}(t_{\rm e})) \nonumber \\
                                                                         & -\tau_{j{\rm e}}^{\rm max}\alpha_{j{\rm e}}(t_{\rm e})\} \label{eq_t}
\end{align}
\begin{align}
  \theta_{j}(t_{\rm e}+\Delta t) & =\frac{1}{K_{j}(\alpha_{j\rm{f}}(t_{\rm{e}}+\Delta t))} \{ \tau_{j{\rm f}}^{\rm max}\alpha_{j{\rm f}}(t_{\rm e}+\Delta t) \nonumber \\
                                 & \quad -\tau_{j{\rm e}}^{\rm max}A_{\rm e}(t_{\rm e}+\Delta t)(1-\alpha_{j{\rm f}}(t_{\rm e})+\Delta t) \}. \label{eq_tD}
\end{align}
Solving the above equations for $A_{\rm{e}}$ and organizing it by substituting in equation (\ref{tauF}), we obtain the following torque $\tau_{j\rm{f}}$ expression, when the flexion direction is determined.
\begin{eqnarray}
  \begin{cases}
    \tau_{j\rm{f}}=C^{+}_{\rm{f}}\tau_{j\rm{f}}^{\rm{max}}\alpha_{j\rm{f}} \\
    \tau_{j\rm{e}}=C^{+}_{\rm{f}}\frac{1-\alpha_{j\rm{f}}}{1-\alpha_{j\rm{f}}^{\rm{post}}}(\tau_{j\rm{f}}^{\rm{max}}\alpha_{j\rm{f}}^{\rm{post}}-K_{j}(\alpha_{j{\rm f}}^{\rm post})\theta_{j\rm{e}}^{\rm{pre}})
    \label{tauF2}
  \end{cases}
  ,
\end{eqnarray}
where $\theta_{j\rm{e}}^{\rm{pre}}=\theta_{j\rm{e}}(t_{\rm{e}})$ is the equilibrium angle just before switching, and $\alpha_{j\rm{f}}^{\rm{post}}=\alpha_{j\rm{f}}(t_{\rm{e}}+\Delta t)$ is the muscle contraction level in the flexion direction immediately after switching.
Similarly, when the model is applied in the flexion direction at time $t_{\rm{f}}$ and there is a switch to the model in the extension direction at time $t_{\rm{f}}+\Delta t$, the torques $\tau_{j\rm{f}}$ and $\tau_{j\rm{e}}$ in the extension direction can be written as follows:
\begin{eqnarray}
  \begin{cases}
    \tau_{j\rm{f}}=C^{+}_{\rm{e}}\frac{1-\alpha_{j\rm{e}}}{1-\alpha_{j\rm{e}}^{\rm{post}}}(\tau_{j\rm{e}}^{\rm{max}}\alpha_{j\rm{e}}^{\rm{post}}+K_{j}(\alpha_{j\rm e}^{\rm post})\theta_{j\rm{f}}^{\rm{pre}}) \\
    \tau_{j\rm{e}}=C^{+}_{\rm{e}}\tau_{j\rm{e}}^{\rm{max}}\alpha_{j\rm{e}}
    \label{tauE2}
  \end{cases}
  ,
\end{eqnarray}
where $\theta_{j\rm{f}}^{\rm{pre}}=\theta_{j\rm{f}}(t_{\rm{f}})$ is the equilibrium angle just before switching, and $\alpha_{j\rm{e}}^{\rm{post}}=\alpha_{j\rm{e}}(t_{\rm{f}}+\Delta t)$ is the muscle contraction level in the extension direction immediately after switching.
Finally, using (\ref{motion}), (\ref{tauF2}), and (\ref{tauE2}), the control model for the direction $i \in \{\mathrm{f}, \mathrm{e}\}$ is obtained as follows:
\begin{align}
    &I_{j}\ddot{\theta_{j}}(t)+B_{j}(\alpha_{ji})\dot{\theta_{j}}(t)+K_{j}(\alpha_{ji})(\theta_{j}(t)-\theta_{ji}^{0}(t))   \nonumber      \\
    &\quad \qquad \qquad \qquad= C^{+}_{i} \delta_i \tau_{ji}^{\rm{max}}(\alpha_{ji}(t)-V_{i}(t)\alpha_{ji}^{\rm{post}}), \\
    &\theta_{ji}^{0}(t) = C_{i}^{+}\frac{K_{j}(\alpha_{ji}^{\rm post})}{K_j(\alpha_{ji})}V_{i}(t)\theta_{ji}^{\rm{pre}}+C_{i}^{-}\theta^{\mathrm{eq}}_{ji}(t-\Delta t),
    \label{modelf}
\end{align}
where $\delta_{\mathrm{f}} = 1$, $\delta_{\mathrm{e}} = -1$, and
\begin{eqnarray}
  V_{i}(t)=\frac{1-\alpha_{ji}(t)}{1-\alpha_{ji}^{\rm{post}}}.
\end{eqnarray}

\section{Simulation Experiments}
\subsection{Methods}
Numerical simulation using artificial data was conducted to study whether the control model satisfies the design specifications.
For simplicity, the number of joints was $j=1$, and the change in the equilibrium angle $\theta^{\mathrm{eq}}$ was confirmed for the two types of input signals.
\\
\textbf{Input 1}
\begin{align}
  \alpha_{\rm{f}} &=
  \begin{cases}
    t/10 & \mathrm{if}\ 0\leq t<5~\rm{s}    \\
    (t - 20) / 10 & \mathrm{if}\ 20\leq t\leq30~\rm{s} \\
    0 & \mathrm{otherwise}
  \end{cases}
  ,\\
  \alpha_{\rm{e}} &=
  \begin{cases}
    (t - 10) / 10 & \mathrm{if}\ 10\leq t<15~\rm{s} \\
    0 & \mathrm{otherwise}
  \end{cases}
  .
\end{align}
\textbf{Input 2}
\begin{align}
  \alpha_{\rm{f}} &= \frac{1}{2}\sin(0.2\pi t-\frac{\pi}{2})+\frac{1}{2}, \\
  \alpha_{\rm{e}} &=
  \begin{cases}
    0 & \mathrm{if}\ 0\leq t<5~\rm{s}\\
    \frac{1}{2}\sin(0.2\pi t-\frac{3}{2}\pi)+\frac{1}{2} & \mathrm{if}\ 5\leq t\leq30~\rm{s}
  \end{cases}
  .
\end{align}
Here, $\alpha_{\rm f}$ and $\alpha_{\rm e}$ were assumed to be the muscle contraction level obtained from the flexor and extensor muscles, respectively.
The threshold $\widetilde{\lambda}_{i}$ was defined as the maximum value of the muscle contraction level in the classification section.
In addition, the section where $\alpha_{\rm f}>\alpha_{\rm e}$ represents the classification section of the flexor muscle while the section where  $\alpha_{\rm f}<\alpha_{\rm e}$ represents the classification section of the extensor muscle.
In this experiment, only the muscle contraction level of the classified muscle was used as the input of the control model.
The impedance parameters were set to the values of the wrist joints (Table~\ref{parameter}) measured in a previous study~\cite{Tsuji2010-cg}.

\begin{table}[t]
  \centering
  \caption{Impedance parameters used in the experiments}
  \begin{threeparttable}
  \begin{tabular}{@{}ccccccccc@{}}
    \toprule
    \multicolumn{9}{c}{Wrist joint}\\
    \midrule
    \multirow{2}{*}{\begin{tabular}[c]{@{}c@{}}Inertia\\$I_j$\end{tabular}} & \multicolumn{3}{c}{Viscosity $B_j(\alpha_j)$} & \multicolumn{3}{c}{Stiffness $K_j(\alpha_j)$} & \multirow{2}{*}{$\tau_{j{\rm f}}^{\rm max}$} & \multirow{2}{*}{$\tau_{j{\rm e}}^{\rm max}$}                                         \\ 
    \cmidrule(l){2-4} \cmidrule(l){5-7}
                                                & $b_{j,1}$                                     & $b_{j,2}$                                     & $b_{j,3}$                                    & $k_{j,1}$                                    & $k_{j,2}$ & $k_{j,3}$ &       &       \\ 
    \midrule
    0.004                                       & 0.14                                          & 0.2                                           & 0.144                                        & 32.8                                         & 0.6       & 3.2       & 46.12 & 44.25 \\ 
    \bottomrule
    \toprule
    \multicolumn{9}{c}{Finger joint}\\
    \midrule
    \multirow{2}{*}{\begin{tabular}[c]{@{}c@{}}Inertia\\$I_j$\end{tabular}} & \multicolumn{3}{c}{Viscosity $B_j(\alpha_j)$} & \multicolumn{3}{c}{Stiffness $K_j(\alpha_j)$} & \multirow{2}{*}{$\tau_{j{\rm f}}^{\rm max}$} & \multirow{2}{*}{$\tau_{j{\rm e}}^{\rm max}$}                                         \\ 
    \cmidrule(l){2-4} \cmidrule(l){5-7}
                                                & $b_{j,1}$                                     & $b_{j,2}$                                     & $b_{j,3}$                                    & $k_{j,1}$                                    & $k_{j,2}$ & $k_{j,3}$ &       &       \\ 
    \midrule
    0.001                                       & 0.08                                          & 0.2                                           & 0.090                                        & 0.90                                         & 0.6       & 0.3       & 0.8 & 0.8 \\ 
    \bottomrule
  \end{tabular}
  \begin{tablenotes}[para,flushright]
		$I_j$: [$\mathrm{kgm}^2$], $b_1, b_3$: [$\mathrm{Nms/rad}$], $k_1, k_3$: [$\mathrm{Nm/rad}$]
	\end{tablenotes}
  \end{threeparttable}
  \label{parameter}
\end{table}

\subsection{Results and Discussion}

\begin{figure}[t]
  \centering
  \includegraphics[width=\hsize]{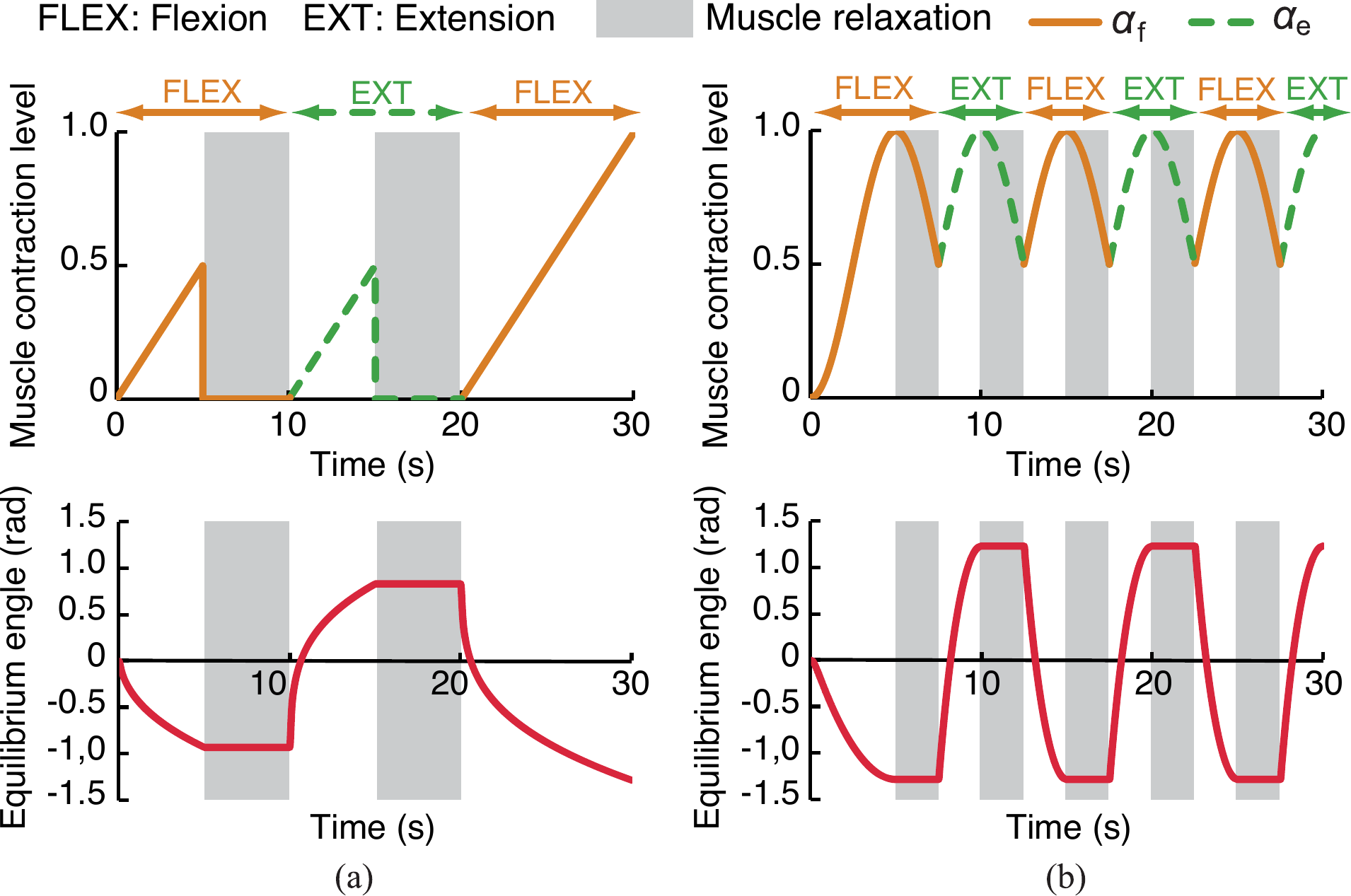}
  \caption{Simulation results of the equilibrium angle $\theta^{\mathrm{eq}}$. (a) Results for Input 1. (b) Results for Input 2. 
  Red solid and blue dashed lines are the muscle contraction levels assumed to be obtained for the flexor muscle, $\alpha_{\rm f}$, and extensor muscle, $\alpha_{\rm e}$, respectively.
  Shaded area represents the section during muscle relaxation.
  }
  \label{Simu}
\end{figure}

Fig.~\ref{Simu} shows the artificially generated muscle contraction level $\alpha_i$ and the equilibrium angle $\theta_0$ of the joint obtained from the numerical simulation for (a) Input 1 and (b) Input 2.
The direction of joint extension is defined as positive, and the shaded area in the figure represents the section of muscle relaxation.

In Fig.~\ref{Simu}, when flexors are classified, the equilibrium angle transitions into a negative direction.
In contrast, when extensors are classified, the equilibrium angle transitions into a positive direction.
When considering the muscle relaxation section, the equilibrium angle was maintained at the same level as before muscle relaxation.
Furthermore, the equilibrium angle changes continuously and smoothly before and after the switching of the motion.
The equilibrium angle presents a periodic waveform following the muscle contraction level, even for inputs where the flexor muscle and extensor muscle are intermittently switched as shown in Fig.~\ref{Simu}(b). 
Hence, the continuity of the equilibrium position is maintained successfully.
These results suggest that the control model satisfies the design specifications and maintains the equilibrium position even when the muscle relaxes.

Furthermore, the equilibrium angle shifts in a curved manner with respect to a linear input (Fig.~\ref{Simu}(a)).
The reason is that the proposed method was constructed based on the impedance model, so that the impedance characteristics of a human are reflected in the transition of the angle.
The simulation results revealed that the proposed control method can replicate the motion characteristics of the muscle based on the muscle model and the motion characteristics of the joint by the impedance control, realizing a behavior approaching that of humans.

\section{EMG-Based Control Experiments}
\subsection{Methods}
To evaluate the applicability of the proposed method to the prosthetic hand control, we constructed a myoelectric prosthetic hand system and performed control experiments using EMG signals.
All measurement experiments were approved by the Hiroshima University Ethics Committee (registration number E-840).

\subsubsection{Experimental System}
\begin{figure}[t]
	\centering
	\includegraphics[width=\hsize]{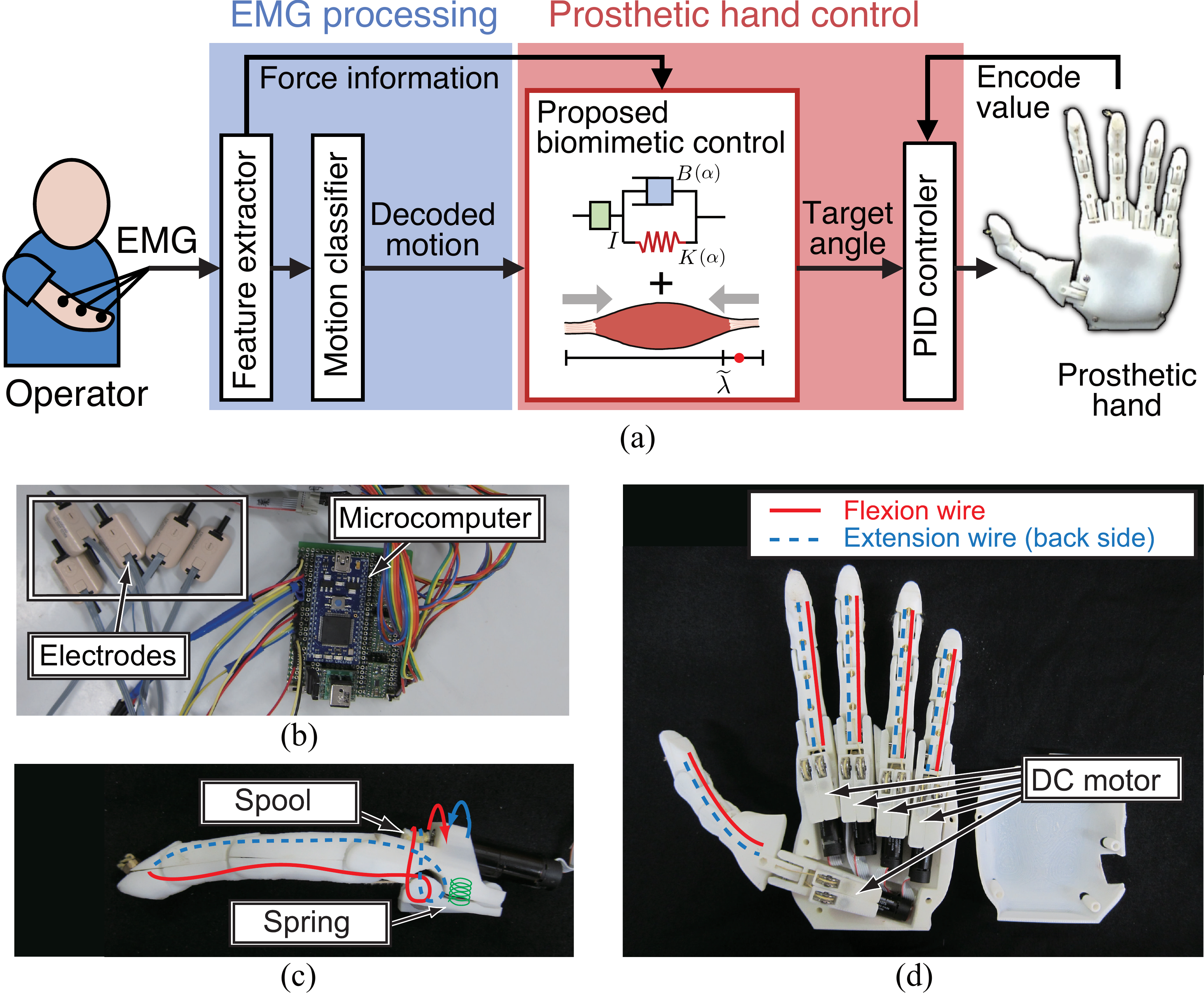}
  \caption{Overview of the experimental prosthetic hand system.
  (a) Based on the measured EMG signals, the force information and EMG feature patterns are calculated.
  The operator's motions are subsequently decoded from the EMG patterns, and the joint angle of the prosthetic hand is calculated using the proposed control method based on the calculated force information and encoded motion. 
  The prosthetic hand movement can then be performed using a PID controller.
  (b--d) The hardware configuration of the system.}
	\label{system}
\end{figure}

Fig.~\ref{system}(a) shows an overview of the control system of the myoelectric prosthetic hand based on the proposed biomimetic control method.
The system consists of three parts: the EMG signal processing, prosthetic hand control, and myoelectric prosthetic hand.
The prosthetic hand movement can be controlled using a proportional-integral-derivative (PID) controller.

The EMG signal recording/processing and the prosthetic hand control were implemented in the electrical apparatus, which consist of bioelectrodes, a microcomputer, and a motor driver (Fig.~\ref{system}(b)). 
We used MyoBock electrodes (OttoBock~13-E200,~Otto Bock HealthCare) for EMG signal recording.
This electrode conducts differential amplification, bandwidth limitation (90--450 Hz), rectification, and smoothing within its internal analog circuit. 
The signal processing part and control system were implemented in a microcomputer (mbed LPC1768).
Motor drivers (Texas Instruments, DRV8835) were used to supply sufficient driving voltage.

The exterior and other parts were printed using a three-dimensional printer.
We designed the parts based on open-source models released by the Open Hand Project~\cite{Gibbard_undated-hd} and Open Bionics Ada Hand~\cite{OpenBionics_undated-dy}.
Each finger has a wire that is wound around a spool (Fig.~\ref{system}(c)).
Subsequently, an actuator based on a DC motor rotates the spool so that the finger flexes.
Thus, the hand can drive each finger independently, providing five degrees of freedom (Fig.~\ref{system}(d)).

The EMG signal processing~\cite{Furui2019-bz} involved feature extraction and motion classification.
The measured EMG signals from the $L$ electrodes were first digitized through A/D conversion. 
The signals were then low-pass filtered, and normalized using pre-measured signals at rest and maximum voluntary contraction.
After that, the EMG pattern was calculated by transforming the normalized signal so that the sum of all channels of the normalized signal was equal to 1.0.
Furthermore, the normalized EMG signal was averaged over all channels to calculate the force information $F_{\mathrm{EMG}}(t)$.
Motion was recognized when $F_{\rm EMG}(t)$ was greater than a threshold $F^{\rm th}$.
Let $m_{\rm f}$ and $m_{\rm e}$ be movements in the flexion and extension directions, respectively, the muscle contraction level in the respective motion directions are defined as follows:
\begin{eqnarray}
  \alpha_{i}(t)=\frac{F_{\rm EMG}(t)}{F^{\rm max}_{m_i = m^{\prime}_i}}~~~~~~(i \in \{{\rm f},~{\rm e}\}),
\end{eqnarray}
where $F^{\rm max}_{m_i}$ is $F_{\rm EMG}(t)$ at maximum muscle contraction with respect to each motion $m_i$~($m_i = 1,~2,~\cdots,~M$;~$M$ is the number of motions) measured in advance.
By treating $\alpha_i(t)$ as the muscle contraction level in the control model, we realize EMG-based voluntary control of the motor angles.
The motion classification uses a recurrent log-linearized Gaussian mixture network (R-LLGMN)~\cite{Tsuji2003-zy}, a recurrent neural network comprising a Gaussian mixture model and a hidden Markov model, thereby considering the time-series characteristics of the operator's motions.
The system handles movements in the flexion and extension directions of each motion as a separate class; hence, the number of classes learned and classified by the R-LLGMN is $M \times 2$.

\subsubsection{Evaluation of the Joint Angle Calculation}
An experiment was conducted on the joint angle calculation for the flexion and extension motion of the wrist to verify the effectiveness of the proposed method.
In the experiment, five healthy male participants (average age: 23.6 years, standard deviation: 1.14 years) adopted a posture with their forearm on a desk and their palm vertical while sitting. 
A cushioning material was placed between the wrist joint and the desk to prevent friction with the desk. 
The EMG signals were measured using two sensors ($L=2$), attached to the flexor carpi radialis and the extensor carpi radialis longus muscles.
Simultaneously, the joint angle of the wrist was measured by a goniometer (SG65, Biometrix), attached from the third metacarpal bone of the back of the hand to the midline of the forearm.
The sampling frequency of the EMG sensors and goniometer was set to 500 Hz.
The experiments considered only the motion of the wrist ($M=1$), and the classes learned and classified using the R-LLGMN were set to two classes of wrist flexion and extension.

First, the participants were instructed to flex and then extend the wrist, and the corresponding EMG signals were acquired.
To train the R-LLGMN, we considered 100 samples randomly selected from EMG patterns for training in flexion and extension.
A target joint angle on the screen was presented to the participants.
They were instructed to flex and extend their wrist joint while following the target joint angle.
The joint angles of the participants were also shown to them in real-time in the form of a line graph.
In the experiment, three trials were performed for each of the two tasks.
Task 1 comprised an initial rest (5 s), wrist flexion of $\pi/3$ rad (5 s), and remain in the same position with muscle relaxation (5 s).
Task 2 comprised an initial rest (3 s), wrist flexion of $\pi/4$ rad (4 s), remain with muscle relaxation (4 s), wrist extension of $7\pi/18$ rad (4 s), and remain with muscle relaxation (4 s).
The state where the forearm and palm are horizontal is assumed to be the initial state (0 rad).
In the experiment, the number of components in the R-LLGMN was set to 1, the threshold was set to $F^{\rm th}=0.02$, and the cut-off frequency of signal filtering was set to $f_{\mathrm{c}}=8.0$ Hz.

The muscle contraction level $\alpha_i$ calculated from the EMG signals and the motion decoded by R-LLGMN were considered the inputs of the proposed method.
We compared the calculated joint angle and the actual joint angle measured by the goniometer to verify the accuracy.
The root-mean-square error (RMSE) between the measured and calculated angles was calculated at each time point as an index of the accuracy with respect to the calculated joint angle.
For comparison purposes, the joint angles were similarly obtained using a conventional impedance control method~\cite{Tsuji2010-cg} and a proportional control method~\cite{OttoBock_undated-wi}.
The RMSE was also calculated.
The impedance parameters were set as in Section 3.1.
In the proportional control method, the calculated joint angle was limited to be between the range of $-\pi/2$ and $7\pi/18$, which corresponds to the motion range of the wrist joint.

\subsubsection{Prosthetic Hand Control}
Control experiments of a myoelectric prosthetic hand were performed using a prosthetic hand system that applies the proposed control method.
The participant was a healthy male (aged 24 years), and the electrode placement and movements to be performed were the same as in the previous experiment.
In this experiment, the flexion and extension of the wrist corresponded to the grasping and opening motions of the prosthetic hand, respectively.
The participant performed flexion and extension of the wrist, in turn, with a muscle relaxation period between each motion.
The D component of the PID control was omitted, and the control cycle of the microcomputer was 5 ms.
The impedance parameters were set to the values of the finger joint (Table~\ref{parameter}) based on a previous study~\cite{Tsuji2010-cg}.

\subsection{Results}
\begin{figure*}[t]
	\centering
	\includegraphics[width=\hsize]{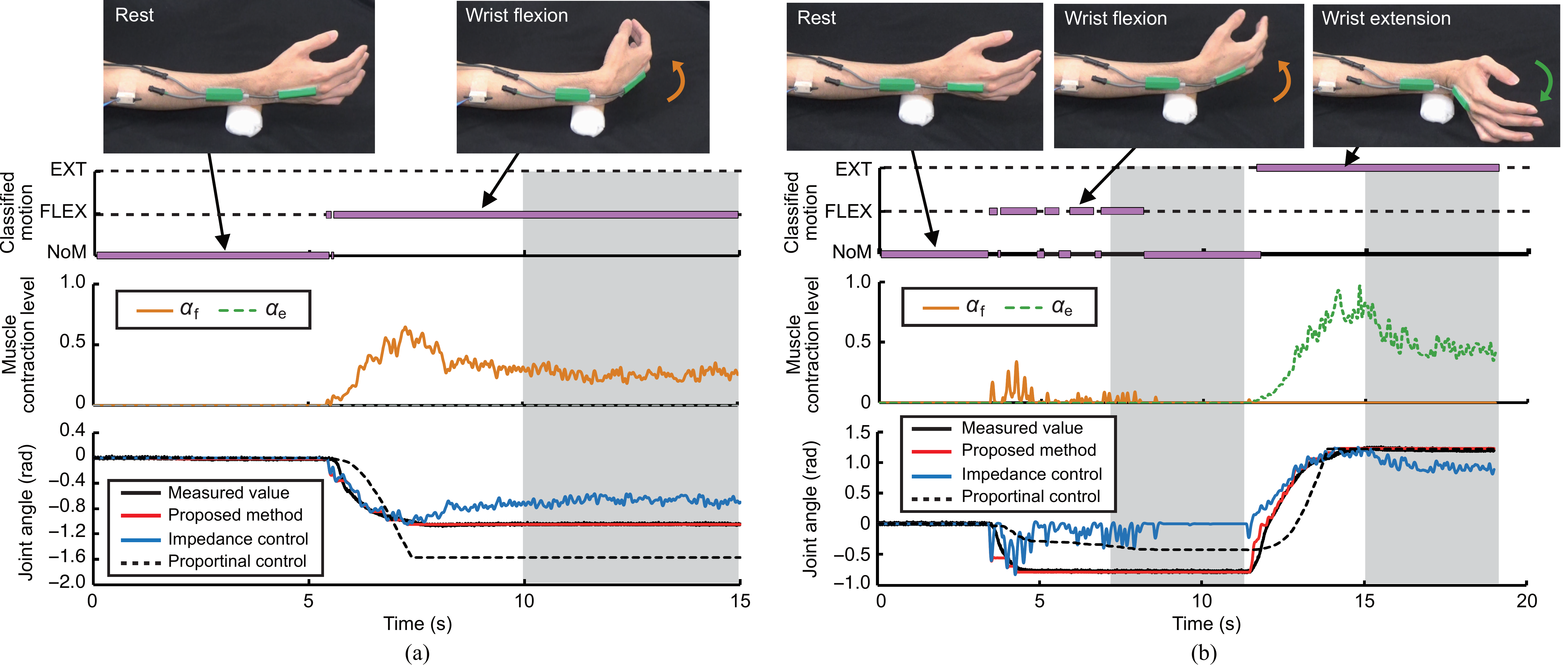}
  \caption{Examples of the measured muscle contraction level and calculated angles. 
  (a) Task 1. (b) Task 2.
  The shaded areas represent the sections of muscle relaxation. 
  The motions conducted by the participants were no motion (NoM), wrist flexion (FLEX), and wrist extension (EXT).}
	\label{task1}
\end{figure*}

\begin{figure}[t]
	\centering
	\includegraphics[width=0.85\hsize]{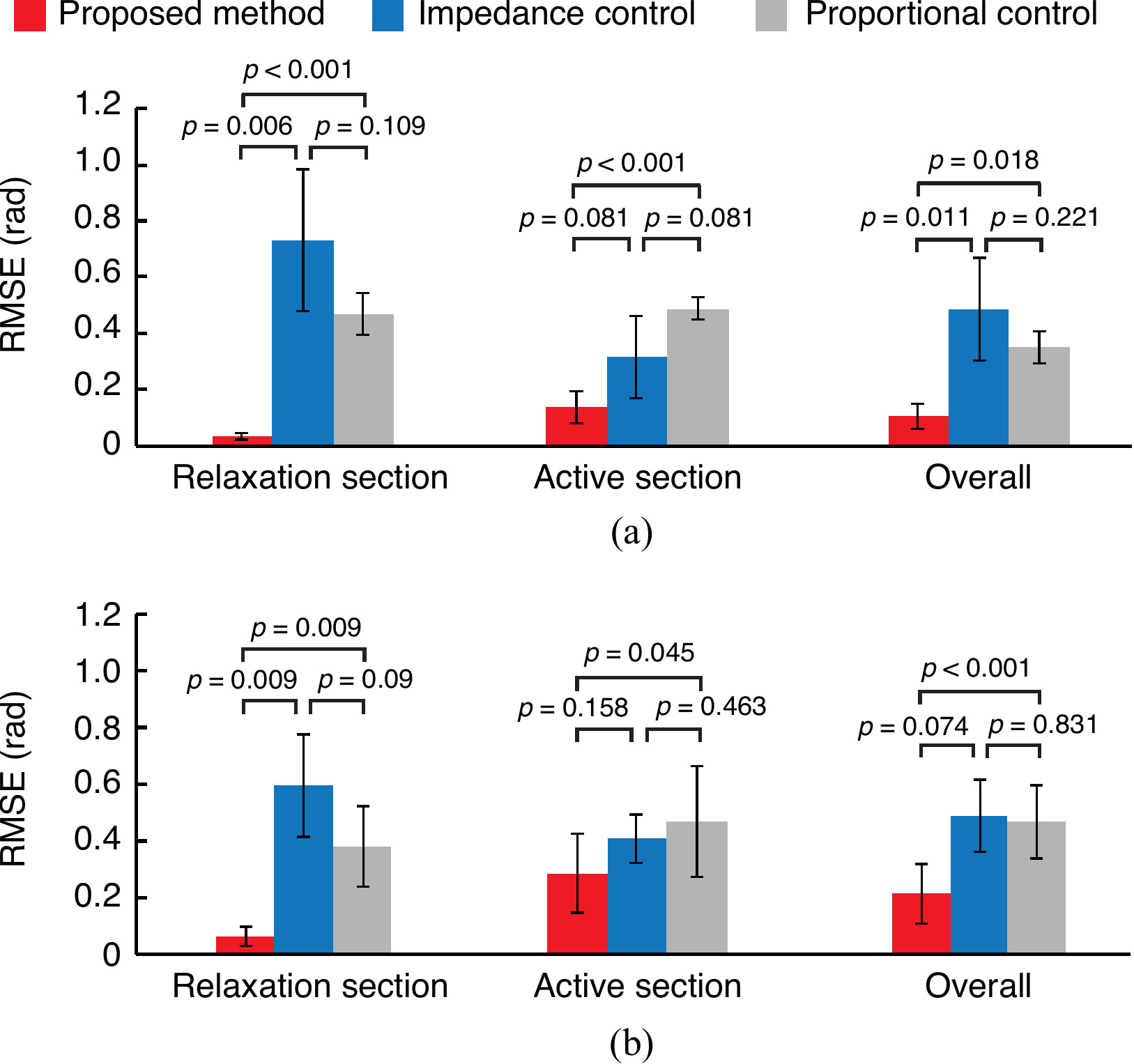}
  \caption{Root-mean-square error (RMSE) for relaxation section, active section, and overall. 
  (a) Taks 1. (b) Task 2. 
  Error bars represent the standard deviations of all participants.
  $p$ values obtained by paired $t$-test with Holm adjustment are also presented.}
	\label{RMSE1}
\end{figure}

Fig.~\ref{task1} shows examples of joint angle calculation for each task.
The upper, middle, and lower panels represent the classified motions, muscle contraction levels, and joint angles, respectively.
The shaded area in the figure corresponds to the section where the participants were instructed to relax.
Fig.~\ref{RMSE1} shows the average RMSE over participants for each task.
The statistical test results of a paired $t$-test (significance level of 5\%) with Holm adjustment are also shown.

\begin{figure}[t]
	\centering
	\includegraphics[width=\hsize]{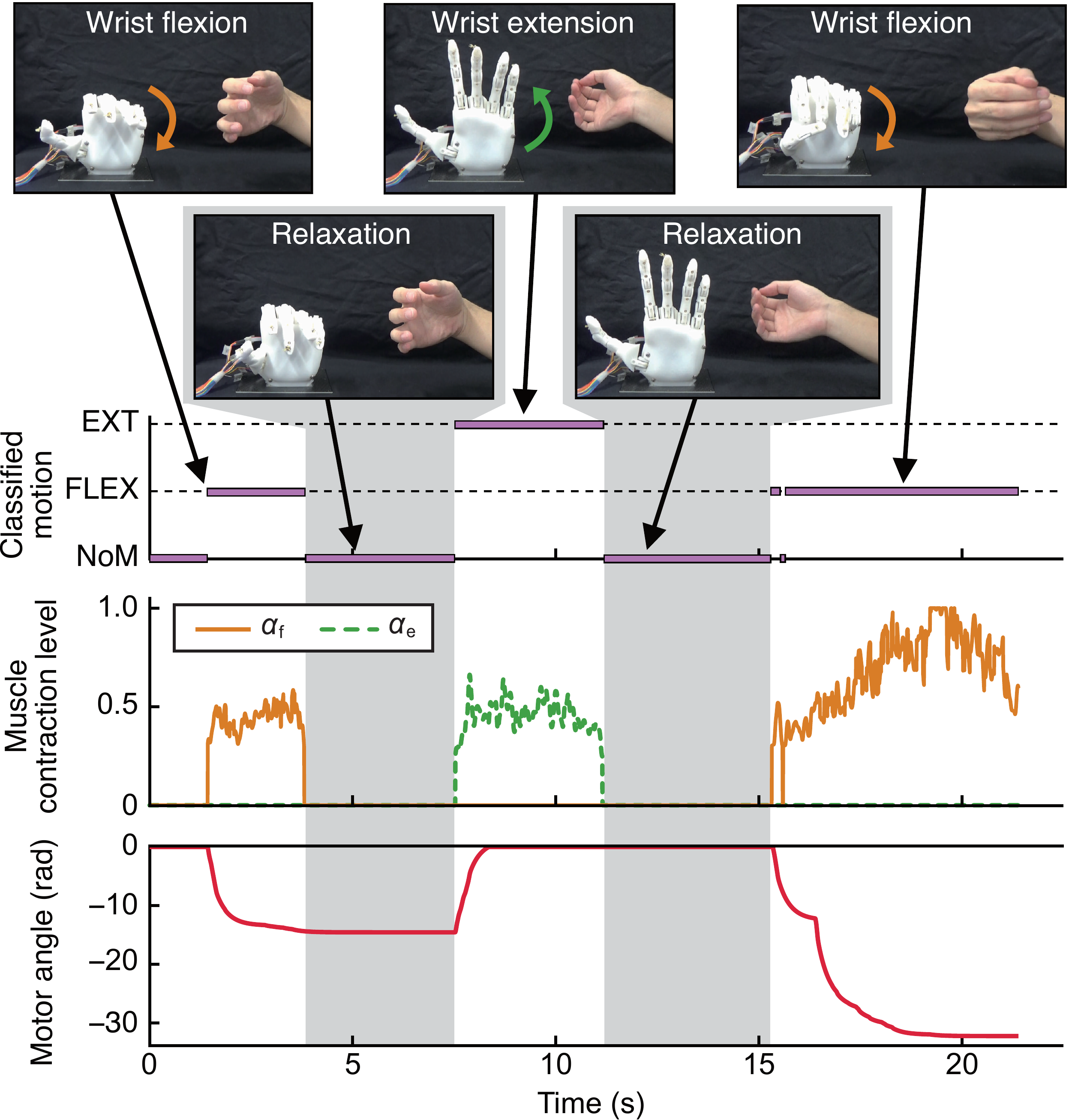}
  \caption{Results of the prosthetic hand control experiment. 
  The shaded areas represent the sections of muscle relaxation. 
  The motions conducted by the participants were no motion (NoM), wrist flexion (FLEX), and wrist extension (EXT).
  The wrist flexion and extension of the participant correspond to the hand closing and opening of the prosthetic hand.}
	\label{Actu}
\end{figure}

Fig.~\ref{Actu} shows the results of the control experiments.
The upper, middle, and lower panels represent the classified motions, muscle contraction levels, and motor output angles, respectively.
The opening motion of the prosthetic hand was set as the initial state, implying that the initial angle was zero.
Shaded areas represent sections of muscle relaxation.

\subsection{Discussion}
In the joint angle calculation experiment, the RMSE was used to evaluate the error between the actual human joint angle and the calculated joint angle based on the control models, using muscle contraction information.
The RMSE of the proposed method was smaller than that of the comparison methods in both tasks, suggesting that the proposed method can realize more human-like wrist motions with relatively high accuracy.
Although this tendency was similar for all sections, it was more remarkable in the relaxation section.
Moreover, the proposed method had significantly smaller RMSEs.

In the conventional impedance control method, as the origin of the joint angle is always fixed to the initial position, the equilibrium and joint angles converge to the initial position when the muscle is relaxed.
Therefore, the error of the impedance control method is higher in the section where the wrist joint angle is maintained in the muscle relaxation.
The proportional control method showed lower RMSEs than the impedance control method in the muscle relaxation section.
The reason is that the angle saturation in the proportional control method, and the RMSE becomes zero when the target angle coincides with the saturation angle, as shown in Fig.~\ref{task1}(b).
In the active sections, the RMSEs of the proportional control method were higher than those of the proposed model and the impedance control method. 
The proportional control method cannot calculate the joint angle by considering the movement characteristics of a human, resulting in errors in the process of transition of the joint angle.
In contrast, the proposed method controls the equilibrium angle based on the muscle model dynamically. 
The error concerning the measured joint angle was small because it was maintained, even when the muscle was relaxed. 
Moreover, as the control model involves the impedance characteristics of the human, the human-like smoothness was replicated in the change of the joint angle; hence, the error during the transition was small. 
These results suggest the effectiveness of the proposed method to dynamically control the equilibrium position, even during muscle relaxation, in reproducing human joint movements.

Furthermore, we verified the applicability of the proposed control method through an experiment considering the control of the prosthetic hand.
In Fig.~\ref{Actu}, the joint angle of the prosthetic hand is smoothly controlled according to the operator's movements.
We can also confirm that the motor angle is maintained during muscle relaxation by dynamically changing the equilibrium position.
Therefore, by using the proposed control method, the state of the muscle, such as contraction or relaxation, was revealed to be reflected in the motion of the prosthetic hand, and that human motion characteristics can be reproduced.

In the conventional impedance control method, maintaining the muscles contracted through invariance of the equilibrium angle is necessary to maintain the joint angle, which may increase the burden on the user.
The EMG signal pattern is known to change due to muscle fatigue when the muscles are contracted for a long time, affecting the operability of myoelectric prosthetic hands~\cite{Asghari_Oskoei2007-nq,Yang2019-xn}.
In contrast, in the proposed method, the equilibrium angles are dynamically controlled, so that the motion can be maintained even when the muscle is relaxed.
Therefore, the proposed method has the potential to reduce muscle fatigue and improve the operability of myoelectric prosthetic hands during long-term use.

\section{Conclusions}
In this paper, a biomimetic control method was proposed based on the $\lambda$-model, which is a muscle motion model.
With the proposed method, it is possible to dynamically and continuously change the equilibrium angle of a joint according to the state of a muscle.
Moreover, it is possible to realize a smooth motion based on the impedance characteristics of a person.

Simulation experiments confirmed that the proposed control method maintains the equilibrium angle even when the muscle is relaxed.
The results showed that continuous and smooth control could be realized.
In the joint angle calculation experiment, we compared the joint angle calculated by the proposed control model, as well as by the conventional control methods, and that measured fro the participants.
The results revealed that the proposed method outperforms the conventional control method in terms of matching actual joint angle changes.
Moreover, the proposed method was introduced into the control system of a prosthetic hand, and the control experiment was conducted by a non-amputee participant.
The results suggested that a prosthetic hand movement similar to human behavior could be realized by using the proposed method, which would lead to the reduction of muscle fatigue during long-term use.

A verification experiment was performed assuming simple muscles such as single flexor and extensor muscles and simple movements such as wrist flexion and extension.
However, practical myoelectric prosthetic hands must realize complex motions, such as the independent flexion and extension of five fingers, with high accuracy.
Therefore, in the future, we will study a motion classification method for each flexor muscle and extensor muscle based on the pattern classification and construct a prosthetic-hand control system that can realize complicated motion.
We also plan to conduct control experiments on amputee patients.

\addtolength{\textheight}{-12cm}   

\end{document}